\pdfoutput=1

\documentclass[11pt]{article}

\usepackage[final]{acl}

\usepackage{times}
\usepackage{latexsym}

\usepackage[T1]{fontenc}

\usepackage[utf8]{inputenc}

\usepackage{microtype}

\usepackage[ampersand]{easylist}
\usepackage{wrapfig, boxedminipage}
\usepackage{amsmath}
\usepackage{amssymb}
\usepackage{xcolor}
\usepackage{colortbl}
\usepackage{graphicx}
\usepackage{subfig}
\usepackage{algorithm}
\usepackage{algpseudocode}
\usepackage[frozencache=true,cachedir=minted-cache]{minted} 
\usepackage{booktabs}
\usepackage{multicol}
\usepackage{multirow}
\usepackage{makecell}

\title{Graph Guided Question Answer Generation for Procedural Question-Answering}

\author{Hai X. Pham\textsuperscript{\rm 1}\thanks{\quad Corresponding author (pham.xuan.hai@outlook.com)} \quad Isma Hadji\textsuperscript{\rm 1} \quad Xinnuo Xu\textsuperscript{\rm 1} \quad Ziedune Degutyte\textsuperscript{\rm 1} \quad Jay Rainey\textsuperscript{\rm 1} \\ {\bf Evangelos Kazakos}\textsuperscript{\rm 1}\thanks{\quad E. Kazakos contributed to this work while at SAIC-C.} \quad {\bf Afsaneh Fazly}\textsuperscript{\rm 2} \quad {\bf Georgios Tzimiropoulos}\textsuperscript{\rm 1} \quad {\bf Brais Martinez}\textsuperscript{\rm 1} \\
        \textsuperscript{\rm 1}Samsung AI Center, Cambridge \quad \quad
        \textsuperscript{\rm 2}Samsung AI Center, Toronto}

\begin{document}
\maketitle

\begin{abstract}
In this paper, we focus on task-specific question answering (QA). To this end, we introduce a method for generating exhaustive and high-quality training data, which allows us to train compact (e.g., run on a mobile device), task-specific QA models that are competitive against GPT variants. The key technological enabler is a novel mechanism for automatic question-answer generation from procedural text which can ingest large amounts of textual instructions and produce exhaustive in-domain QA training data. While current QA data generation methods can produce well-formed and varied data, their non-exhaustive nature is sub-optimal for training a QA model. In contrast, we leverage the highly structured aspect of procedural text and represent each step and the overall flow of the procedure as graphs. We then condition on graph nodes to automatically 
generate QA pairs in an exhaustive and controllable manner. Comprehensive evaluations of our method show that: 1) small models trained with our data achieve excellent performance on the target QA task, even exceeding that of GPT3 and ChatGPT despite being several orders of magnitude smaller. 2) semantic coverage is the key indicator for downstream QA performance. Crucially, while large language models excel at syntactic diversity, this does not necessarily result in improvements on the end QA model. In contrast, the higher semantic coverage provided by our method is critical for QA performance.
\end{abstract}

\setlength{\belowcaptionskip}{-3pt}

\linespread{0.7}

\section{Introduction}
\label{sec:introduction}

Asking questions is a natural way for humans to understand how to perform a task. Questions that pertain to a given procedure (i.e., a structured task such as cooking a recipe) encompass both factual questions about a given step (e.g., what tools are used in a given step), as well as questions that span across multiple steps (e.g., the order of steps). A smart AI agent should be able to handle both types of questions to assist humans.

While GPT models and competing alternatives have shown impressive results on multiple applications, including QA, they require large amounts of cloud computing resources due to their extreme sizes, thus being inviable as the QA models behind a smart assistant. We show that it is possible to train \textit{task-specific} small models (e.g. suitable for running on a mobile phone), that at the same time are as accurate and complete as GPT variants on the target task. 

High-quality in-domain training data is however required but, unfortunately, most QA datasets focus on general text comprehension where the answers can be spans from the text \citep{RajpurkarEtal_2016_Squad_D16-1264, Dunn2017SearchQAAN, JoshiEtal_2017_Triviaqa_P17-1147,YangEtal_2018_Hotpotqa_D18-1259}, free-style answers about a specific context \citep{DBLP:conf/nips/NguyenRSGTMD16, he-etal-2018-dureader} or obtained from a conversation history in conversational QA \cite{reddy-etal-2019-coqa}. Similarly, collecting high-quality QA data at scale requires expensive labeling efforts. This motivates our paper, where we propose a method that ingests large quantities of procedural instructions (e.g. cooking recipes) and automatically generates extensive Procedural QA (PQA) training pairs that can be used to fine-tune a well-performing small language model.

In particular, the goal is to automatically generate PQA pairs that elicit information both from single sentences (or steps) in a procedure as well as information that requires reasoning over multiple steps to understand the temporal aspect of a procedure.
While there have been efforts to create multi-modal QA datasets from recipes that require alignment between vision and text \cite{yagcioglu2018recipeqa,pustejovsky2021designing}, to the best of our knowledge, our work is the first that specifically concentrates on extracting a rich set of QA pairs from procedural text.
We focus on cooking recipes as a type of procedural text. In a cooking scenario, single sentence-based questions span local concepts (e.g., quantities of ingredients, cooking times, and tools), while temporal questions cover multiple steps (e.g., order of actions, the contents of mixtures at certain steps). 

To tackle this problem, we propose to leverage the highly-structured nature of procedural text and represent the semantics of the procedure as graphs from which we can automatically generate PQA pairs. Specifically, to cover all question types pertaining to individual steps in a recipe, we rely on Abstract Meaning Representation (AMR) graphs \cite{Banarescu2013-oy}. 
We perform a controlled set of transformations on the AMR graph of a step to generate a number of question AMRs and then generate questions from those AMRs using a pre-trained AMR-to-text model.
For temporal questions that span across multiple steps, we start by converting the recipe into an action flow graph \citep{momouchi-1980-control,hamada-2000-structural,YamakataEtal_2020_English_2020.lrec-1.638,graph2vid_eccv22} using a neural graph parser \citep{DonatelliEtal_2021_Aligning_2021.emnlp-main.554}. 
We then extract all potential temporal answers by traversing the graph, and
generating temporal question templates in the AMR space. We then, once again, rely on AMR-to-text models to generate corresponding questions.\footnote{We use action flow graphs rather than more recent work on multi-sentence DocAMR \cite{naseem2022docamr} as DocAMR does not consider the temporal nature of procedural text.} 
Optionally, our approach can take advantage of LLMs (e.g., GPT3~\cite{gpt3_neurips20}) to increase the syntactic diversity and semantic coverage of the generated questions, either by improving the wording and paraphrasing the generated questions or via directly replacing the graph-to-text generation model with a GPT-based solution that relies on content selected with our graph-guided approach.

Extrinsic evaluation shows the usefulness of our generated data in training question-answering models, which outperforms all considered baselines. 
Our results highlight the importance of devising an approach dedicated to generating QA pairs from procedural text, as we show small models (e.g., T5-base with around 220M parameters) can compete with GPT3 and ChatGPT (175B params) when finetuned on specialized high-quality data.
In addition, intrinsic evaluation of our generated data demonstrates its superiority in terms of diversity, coverage, and overall quality, compared to data generated using several baselines including GPT3-based methods.
\paragraph{Contributions.~} In summary the contributions of our paper are threefold:

\begin{itemize}
    \item We tackle the problem of task-specific QA from procedural text and show that small models can compete with strong LLM baselines when provided with high-quality and exhaustive training data.
    
    \item We introduce a novel graph-based method for question-answer generation from procedural text. 
    We draw on existing graph semantic formalisms, such as Abstract Meaning Representations (AMRs), and also take advantage of Action Flow graphs to represent the temporal relations among recipe steps. This allows us to rely on existing text-to-graph parsers as well as graph-to-text generative models, alleviating the need for specialized annotations. Notably, we also show that our method can take advantage of pre-trained LLMs to increase syntactic diversity and semantic coverage.
    \item We empirically show that our generated QA pairs can be used to train compact question-answering models (e.g., 60M or 220M parameters) that can compete with strong GPT-based baselines. Additionally, we show that the proposed method results in QA pairs with great diversity and high coverage (compared to human-generated question-answer pairs).
\end{itemize}

\section{Related Work}

Question generation is an important topic within the natural language generation community \cite{rus-etal-2010-first}, where given a source text (i.e., context) and a target answer, the task is to generate the corresponding question. The answer is either provided \cite{Song2017AUQ, Zhou2017NeuralQG,ZhaoEtal_2018_Paragraph_D18-1424,ChaiWan_2020_Learning_2020.acl-main.21,chan-fan-2019-recurrent,Wang2020NeuralQG} or automatically extracted from the context \cite{GolubEtal_2017_Two_D17-1087, ScialomEtal_2019_Self_P19-1604, PyatkinEtal_2021_Asking_2021.emnlp-main.108}. Our work follows the latter approach, where we automatically extract answers and generate corresponding questions. 

Existing methods either rely on hand-crafted rules and templates \cite{HeilmanSmith_2010_Good_N10-1086,DBLP:journals/corr/abs-2105-10023,FabbriEtal_2020_Template_2020.acl-main.413,pustejovsky2021designing}, or use annotated data (in the form of text spans as answers, along with corresponding ground-truth questions) to learn to automatically generate QA pairs~\cite{t5-qg,gong2022diffuseq,GolubEtal_2017_Two_D17-1087, ScialomEtal_2019_Self_P19-1604, PyatkinEtal_2021_Asking_2021.emnlp-main.108}.
%
%
Rule-based methods offer more control over the generated data, but are not easily scalable. Learning-based methods offer better scalability, but require costly annotations. Our graph-based approach combines the benefits of the two while addressing their short-comings. Specifically, our graph-guided content selection offers the desired control over the content extracted from procedural text (to form the answer), and draws on generic models for graph-to-text generation to generate questions \cite{amrlib}.

Recent work has shown that Large Language Models (LLMs) such as GPT3 can be used for generating QA pairs based solely on the input context texts \cite{wang2021want, yuan2022selecting}. However, these models are sensitive to the prompt used to generate the data, offer less control over the generated QA pairs, and are not cost-effective. In contrast, we provide evidence that our graph-controlled QA generation approach yields high-quality and diverse data that can be used for training question answering models that compete with LLMs while being orders of magnitude smaller.

\section{Methodology}

In this section, we present our approach to QA generation from procedural text. We introduce the AMR and flow graphs that our method relies on (\S \ref{sec:preli}), and detail our method for generating questions from single instructions (\S \ref{sec:qgen_sent}) and temporal questions spanning across multiple instructions (\S \ref{sec:qgen_temporal}). Lastly, we propose to use LLMs to improve the language quality of generated questions (\S \ref{sec:llm-aug}).

\subsection{Preliminaries}
\label{sec:preli}

\paragraph{Abstract Meaning Representation (AMR):} The AMR abstracts away the syntactic idiosyncrasies of language and instead draws out the logical meaning of text entities and their relations in a sentence, following the conventions of common framesets \cite{Banarescu2013-oy}. 
Figure~\ref{fig:amr_example} depicts a recipe instruction (see caption) and its AMR graph, generated by a text-to-AMR parser \cite{amrlib}, in PENMAN notation. 
As can be seen in this example, the AMR graph specifies all entities (ingredients, tools, cooking time) and their relations (location, duration, manner) in a sentence. We draw on this representation to \emph{exhaustively} identify contents to ask questions about for each individual step in a recipe (see \S \ref{sec:qgen_sent}).


\begin{figure}[!ht]
    \centering
    \begin{minipage}{15em}
    \begin{small}
    (c / cook-01
    \\\hspace*{2em}  :mode imperative
    \\\hspace*{2em}  :ARG0 (y / you)
    \\\hspace*{2em}  :ARG1 (a / and
    \\\hspace*{4em}        :op1 (c2 / chicken)
    \\\hspace*{4em}        :op2 (ii / ingredient
    \\\hspace*{6em}              :mod (o / other)))
    \\\hspace*{2em}  :location (p / pot)
    \\\hspace*{2em}  :duration (t / temporal-quantity
    \\\hspace*{4em}        :quant 20
    \\\hspace*{4em}        :unit (m / minute))
    \\\hspace*{2em}  :manner (h / heat-01
    \\\hspace*{4em}        :mod (m2 / medium))
    \\\hspace*{2em}  :purpose (p2 / prepare-01
    \\\hspace*{4em}        :ARG0 y
    \\\hspace*{4em}        :ARG1 (s / soup)))
    \end{small}
    \end{minipage}
    \caption{\textbf{AMR example.} Linearized AMR graph of the sentence \textit{"Cook chicken and other ingredients in the pot over medium heat for 20 minutes to prepare the soup"}.}
    \label{fig:amr_example}
    \vspace{-5pt}
\end{figure}

\begin{figure*}[ht]
    \centering
    \subfloat[``Shepherd pie'' recipe]{\includegraphics[width=0.38\textwidth]{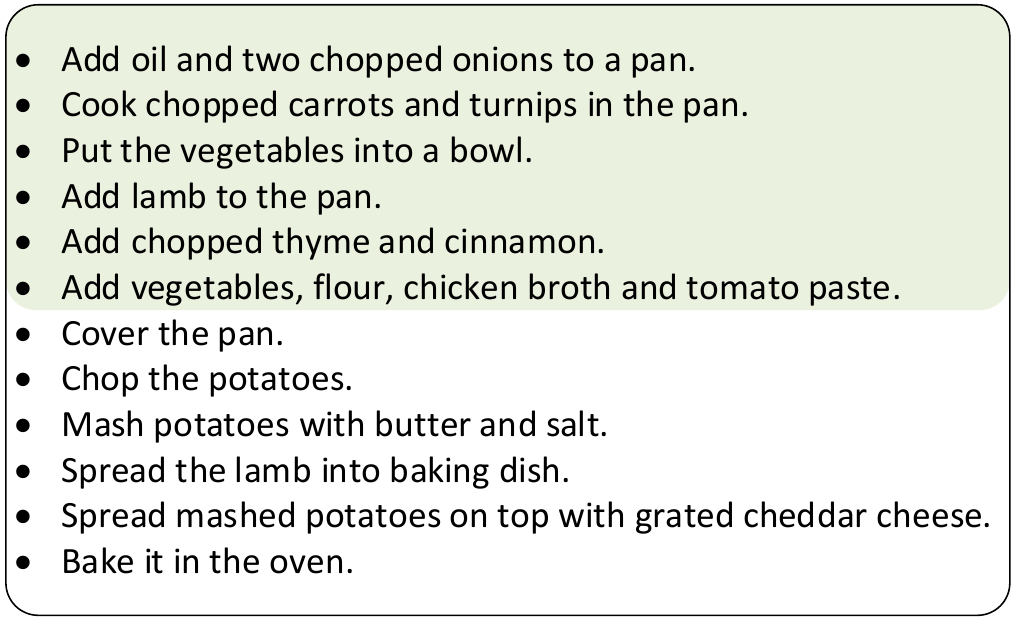}\label{fig:recipe}} \hspace{0.5em}
    \subfloat[Semantic flow graph]{\includegraphics[width=0.55\textwidth]{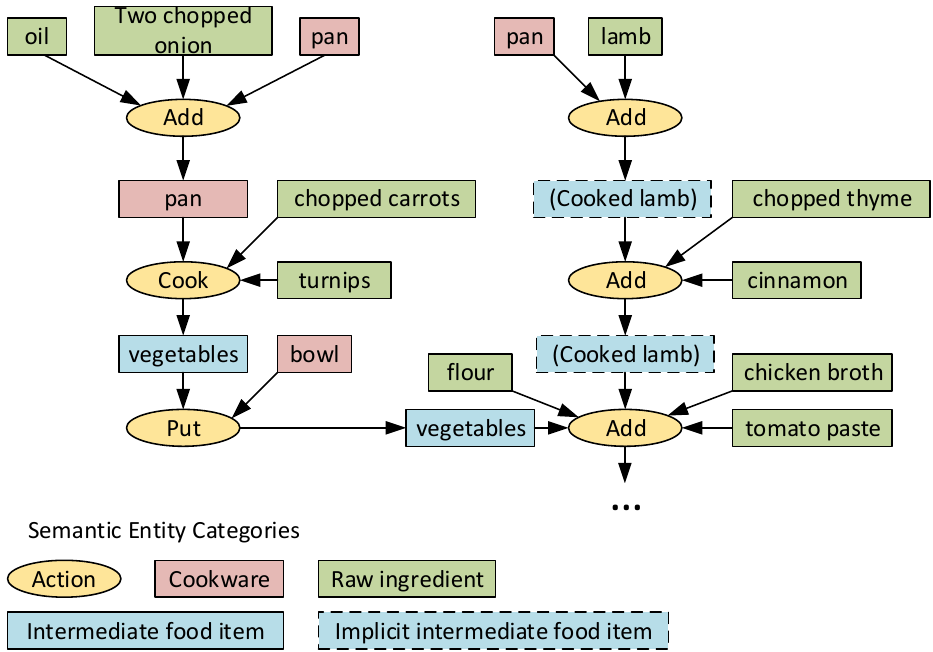}\label{fig:action_graph}}
    \caption{\textbf{Flow graph example}. The action flow (sub-)graph of the highlighted text section in (a) is shown in (b) where word tokens are grouped together to form complete semantic entities belonging to one of the main categories. The semantic graph is further augmented with \textit{implicit entities} to represent entities that are omitted from the text.}
    \label{fig:flow_graph_example}
\end{figure*}

\paragraph{Flow Graphs:} A flow graph \citep{momouchi-1980-control,hamada-2000-structural} is a directed acyclic graph containing actions, objects, other auxiliary entities (nodes) and their relations (edges), which provide essential information to complete a task. Importantly, flow graph relations encode the temporal order of actions and transformations (modifications/combinations) of objects. 
%
We draw on recent flow graph corpora and parsers \citep{YamakataEtal_2020_English_2020.lrec-1.638,DonatelliEtal_2021_Aligning_2021.emnlp-main.554} to generate action flow graphs such as the one shown in Fig. \ref{fig:flow_graph_example}. We then use these graphs to generate questions that require understanding the temporal order of actions and object transformations over time
(see \S \ref{sec:qgen_temporal}).


\subsection{Question generation from a single instruction}
\label{sec:qgen_sent}

We extract the AMR graph for each sentence independently and generate three types of QA pairs from the graph; namely, role-specific, instruction-level, and polarity.

\paragraph{Role-specific QA.} 
We begin by selecting the content that will serve as an answer from the AMR graph. The AMR graph consists of core and non-core roles. We select two main core roles (i.e., \textit{:ARG1}, \textit{:ARG2}) and several non-core roles (i.e., \textit{:time}, \textit{:duration}, \textit{:location}, \textit{:instrument}, \textit{:mod}, \textit{:domain}, \textit{:purpose}, \textit{:accompanier}, \textit{:degree}, \textit{:value} and \textit{:quant}) to generate answers. Each role in the AMR consists of either a single concept (e.g., \textit{:location} in Fig.~\ref{fig:amr_example}) or a subgraph (e.g., \textit{:ARG1}). For roles associated with a single concept, the concepts are used directly as our target answers, whilst for the latter, we use a pre-trained AMR-to-text model \cite{amrlib} to convert the subgraph into a target answer.

To generate questions for each of the selected answers, we construct a corresponding \emph{question} AMR. This is achieved by replacing the answer subgraph in the original AMR with the \textit{amr-unknown} concept and transforming it into a proper AMR graph for natural question generation. The question is then generated using a graph-to-question model finetuned on generic question datasets \citep{RajpurkarEtal_2016_Squad_D16-1264,r2vq_paper}. The transformation algorithms for different roles as well as the graph-to-question generative model training are detailed in the appendix. Figure \ref{fig:amr_q_example} shows examples of questions generated for different roles.
 
\begin{figure*}[ht]
    \centering
    \includegraphics[width=\textwidth]{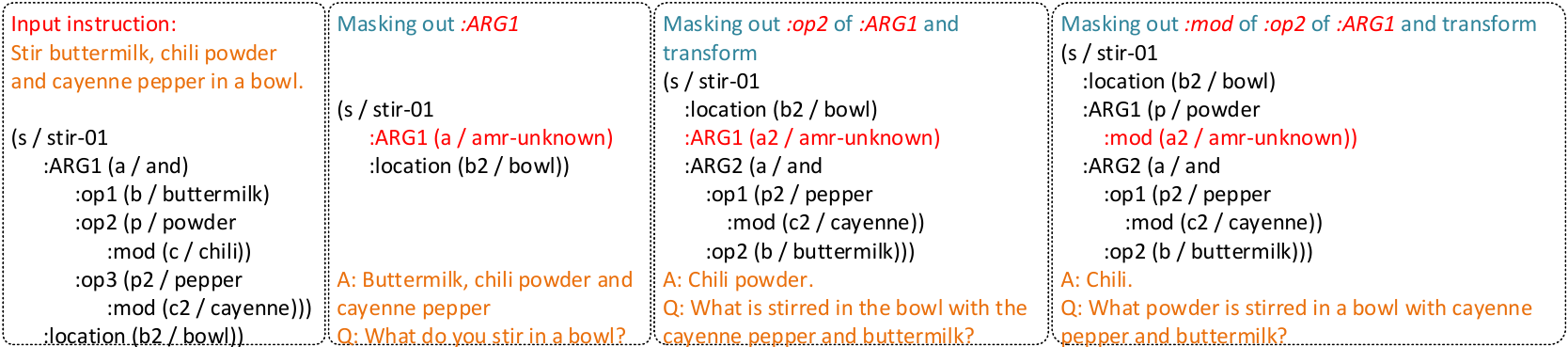}
    \caption{\textbf{Role-specific QA.} Three questions are created by targeting different roles in the input AMR.}
    \label{fig:amr_q_example}
    \vspace{-10pt}
\end{figure*}

\paragraph{Instruction-level QA.}
Instruction-level questions are those for which the answer is the entire sentence. For example, given the instruction [\textit{Slice the onion and coat in flour}] and the question [\textit{How do I prepare the onion?}], the answer is the full instruction. In this category, we cover two types of questions: 1) ``How do you [\textit{do something}]?'' and 2) ``What do we do with [\textit{something}]?''. For the first type, the question AMR is created by adding a \textit{:manner} role with \textit{amr-unknown} concept to the original AMR. 
The second type requires transforming the original AMR into a new AMR in which all core roles (\textit{:ARGx}) are grouped together into \textit{:ARG2} to form [\textit{something}], and the concept of \textit{:ARG1} becomes \textit{amr-unknown}. Once these transformations are applied we again use AMR-to-text model to generate the questions.

\paragraph{Polarity ``yes/no'' QA.} To generate questions with a ``Yes'' answer, we add a new node with the concept \textit{amr-unknown} connected to the main verb root node with the \textit{:polarity} role. 
To generate a question with a ``No'' answer from the same sentence, we further modify the resulting polarity question AMR by replacing one randomly chosen subgraph with a subgraph of the same semantic role sampled from another AMR.

\subsection{Temporal question generation}
\label{sec:qgen_temporal}

We are also interested in questions about the transformation / composition of entities across time, as well as the temporal order of actions. These questions require content selection from multiple steps. We focus on three common types of temporal questions: 1) Composition of a mixture. For example in Fig.~\ref{fig:flow_graph_example}, one may ask about the ingredients that go into \textit{vegetables}; 2) Next or preceding action. For example, in Fig.~\ref{fig:flow_graph_example}, one may want to know \textit{what to do after putting vegetables into a bowl}. Note that in this case the fourth instruction is the start of another subtask, and the correct answer is the sixth instruction. The correct answer is only clear given the action flow graph; 3) The order of actions, e.g., \textit{Is action A performed before/after action B?}.

To cover these question types, we propose a hybrid approach that relies on flow graphs for content selection and AMRs for question generation. More specifically, we create AMR-based templates for each of the question types, and traverse the flow graph to select answer contents (represented as AMR subgraphs extracted from related sentences) for each question type to fill in the templates. Finally, we use the AMR-to-text model to generate the questions. 
We adopt this hybrid strategy for two main reasons. First, temporal questions cannot be constructed directly by modifying the flow graph as was done for the sentence-based QA generation. Second, by composing the questions as AMR graphs, we can rely on the pre-trained AMR-to-text model to generate natural language questions. We now detail the approach adopted for each temporal question type.

\paragraph{Composition of mixtures QA.} We design 12 question templates, represented as AMR graphs, that involve a [\textit{mixture\_name}], such as \textit{``What are the ingredients of the \textbf{[mixture\_name]}?''} 
In the example provided in Fig.~\ref{fig:action_graph}, mixture entities are indicated by cyan boxes in the flow graph. We only apply these templates on \textit{named mixtures}, i.e. we ignore implicit items (dashed boxes in Fig.~\ref{fig:action_graph}) because it is not straightforward to assign names to such references. We then generate questions for each named mixture and traverse the flow graph to obtain the corresponding answer.

\paragraph{Next or preceding actions QA.} We use a question AMR graph template equivalent to \textit{``What do we do after/before \textbf{[action $A_{j}$]}?''} to generate questions. Then, at a particular action step $A_{j}$, the answer is either the next action in the flow graph, $A_k, k=\text{next}(j)$ for a ``next'' question, or the previous action in graph, $A_i, i=\text{pred}(j)$, for a ``before'' question. 
For example, in Fig. \ref{fig:recipe} given that $A_{j=7}$ is \textit{``Cover the pan''}, the preceding action is $A_{i=6}$, \textit{``Add vegetables, flour, chicken broth and tomato paste''}.


\paragraph{Order of actions QA.} Here, we adopt two templates: \textit{``Do we do A or do we do B first?''} that uses AMR \textit{``or''} composition frame; or \textit{``Doing A or doing B, which is first?''} which uses AMR \textit{``amr-choice''} composition frame. We also swap A \& B, so that for each pair of $\{A, B\}$ we can generate four questions. Once again, the answer is directly obtained by traversing the flow graph.

\subsection{QA augmentation with LLMs}
\label{sec:llm-aug}

While the proposed graph-guided method offers a controlled solution to generate diverse QA pairs from procedural text with wide semantic coverage, it can nevertheless still benefit from strong LLMs. In particular, the language used in the QA pairs generated by the proposed method is tightly bound to the language used in the associated recipes. In contrast, humans tend to draw from their own vocabulary when posing questions. Thus, we also introduce two alternative methods to increase the syntactic diversity of the generated QA pairs using LLMs. We use the state-of-the-art GPT3 model.

\paragraph{Answer-based augmentation.}
As one of the strengths of our graph-guided approach is exhaustive content selection, one way to augment it with LLMs is to use answers generated with our approach and rely on GPT3 to generate corresponding questions (prompt details are provided in the appendix). However, we noticed that the questions generated with this approach sometimes do not semantically match the input answers. Therefore, we filter out the generated questions via round-trip consistency similar to \cite{alberti-etal-2019-synthetic}. In particular, we ask GPT3 to generate corresponding answers to the questions which it generated previously. We then compare the answer generated by GPT3 to the original answer in terms of ROUGE-1 metric and only keep the GPT3-based QA pair if the score is $>0.25$.

\paragraph{Paraphrasing-based augmentation.}
Another way to take advantage of LLMs ability to generate diverse syntax is to directly task GPT3 with paraphrasing our graph-guided synthetic questions. Specifically, we paraphrase each question $5$ times and filter out any duplicate questions.
\section{Evaluation}
\label{sec:evaluation}


We describe our experimental setup and question-answer generation baselines in \S \ref{sec:exp-setup}, where we also introduce a new human-annotated PQA set (reference set) used for evaluation. 
\S \ref{sec:QA} shows extrinsic evaluations, where question-answering models are trained with the generated data and evaluated on the aforementioned reference set.
\S \ref{sec:div-cov} describes intrinsic evaluations, assessing the distributions and quality of our generated QA pairs, which further amplifies the significance of our QA generation approach.

\subsection{Experimental setup}
\label{sec:exp-setup}

\paragraph{Data.} We use cooking recipes as a source of procedural text and randomly select 100 recipes from a popular food website \cite{bbcfood}. We use $70$ recipes to generate QA pairs with our method and with the baselines. For evaluation purposes, we compile human-annotated QA pairs from the remaining $30$ recipes and collected $\sim50$ human-generated QA pairs per recipe. This yielded a PQA test set with $1857$ QA pairs, where $\sim30\%$ cover temporal questions that require reasoning over multiple steps in the recipe. We provide more details on the human data collection setup in the appendix.

\paragraph{Baselines.}
We propose a hybrid method for generating PQA pairs, relying on both graph-based logical rules, and trained deep generative models. We thus compare our work to state-of-the-art methods from each type of approach. Specifically, we compare to two rule-based methods \cite{PyatkinEtal_2021_Asking_2021.emnlp-main.108,FabbriEtal_2020_Template_2020.acl-main.413}, and two learning-based methods for QG, including a T5-based model~\citep{t5-qg} finetuned on SQuAD \cite{RajpurkarEtal_2016_Squad_D16-1264} and a diffusion-based model~\cite{gong2022diffuseq}. We also include comparisons to state-of-the-art LLMs. In a preliminary study, we found that GPT3 outperforms ChatGPT for the task of QA data generation so we use the former as a baseline. We consider two strategies, GPT3-sentence and GPT3-recipe, to generate QA pairs from sentences and entire recipes respectively. When evaluating the methods, including ours, we only consider questions that can be answered \emph{solely} from the given context (i.e., recipe). A detailed description of all baselines considered is in the appendix.

\subsection{Extrinsic evaluation}
\label{sec:QA}

\begin{table*}[ht]
    \centering
    \small
    \begin{tabular}{ c|l|c c c c}
    \toprule
    &Generation method & BLEU & F1 & ROUGE-L & BLEURT  \\
    \cmidrule{2-6}
    &SQuADv2 & 6.2 & 23.8  & 23.4  & 34.0 \\
    \midrule
    \multirow{2}{*}{Rule-based} &Role-based QG \citep{PyatkinEtal_2021_Asking_2021.emnlp-main.108} & 3.7  & 20.4  & 19.0  & 37.1 \\
    &Template-based QG \citep{FabbriEtal_2020_Template_2020.acl-main.413} & 6.8  & 25.4  & 25.1  & 35.7 \\
    \midrule
    \multirow{2}{*}{Learning-based} & Diffusion-based QG \citep{gong2022diffuseq} & 2.8  & 17.8   & 16.5  & 35.4 \\
    & T5-based QG \citep{t5-qg} & 5.9  & 27.6  & 27.0  & 38.6 \\
    \midrule
    \multirow{2}{*}{GPT-based} & GPT3-sentence & 3.4  & 18.3  & 17.5  & 32.0 \\
    & GPT3-recipe & 7.1  & 26.6  & 26.2  & 36.7 \\
    \midrule
    \multirow{3}{*}{Ours} & w/o augmentations & 7.2 & 31.4  & 30.7  & 40.0 \\
    & w/ paraphrasing augmentation & {7.3} & {33.5}  & {32.6}  & {42.0} \\
    & w/ answer-based augmentation & {9.9}  & {35.2}  & {34.0}  & {45.8} \\
    \bottomrule
    \end{tabular}
    \caption{\textbf{QA performance for different training data generation approaches}. A \textbf{T5-small} model was fine-tuned on different synthetic QA datasets and test results are computed on the human-annotated reference set.}
    \label{tab:qa_short}
\end{table*}

\begin{table*}[ht]
    \centering
    \small
    \begin{tabular}{c|c| c c c c}
    \toprule
    Model & \#params (B) & BLEU & F1 & ROUGE-L & BLEURT  \\
    \midrule
    T5-small & 0.06  & 7.2 & 31.4  & 30.7  & 40.0 \\
    T5-base & 0.22  & 11.5  & 42.8  & 42.0  & 47.9 \\
    T5-large & 0.77 & 13.9  & 45.0  & 44.0  & 47.9 \\
    \cmidrule(lr){2-2}
    T5-3B & \multirow{4}{*}{3}  & 16.7 & 45.5  & 44.7  & 51.2 \\
    FLAN-T5-XL &  & 16.9 & 49.3 & 48.5 & 51.4 \\
    FLAN-T5-XL \textit{(wP)} &  & 21.8 & 54.0 & 53.1 & 56.3 \\
    FLAN-T5-XL \textit{(wA)} &  & 17.7 & 46.1 & 45.0 & 50.9 \\
    \midrule
    GPT-3 & \multirow{2}{*}{175} & 16.3  & 42.1  & 41.8  & 55.9 \\
    ChatGPT &  & 17.3  & 41.6  & 41.0  & 56.2 \\
    \bottomrule
    \end{tabular}
    \caption{\textbf{QA performance for different model sizes and LLMs.} Different-sized T5 models, trained on our generated data \textit{without LLM augmentations} unless explicitly mentioned: \textit{(wP)} = w/ paraphrasing augmentation, \textit{(wA)} = w/ answer-based augmentation. GPT3 and ChatGPT are the upper bound of the ``generalist'' QA approach. Performance is measured on the human-annotated reference set.}
    \label{tab:model_abla}
    \vspace{-10pt}
\end{table*}

\paragraph{Question answering.} We evaluate the usefulness of the generated data on the important application of question answering. We generate PQA pairs from the 70 recipes in the training set and use them to train a model for question answering. In particular, we target the application of open-ended QA, where given a question, $q$, and corresponding context, $c$, (i.e., recipe in this case), the goal is to generate the correct answer $a=\mathcal{F}(q,c)$. Here, $\mathcal{F}$ is a sequence-to-sequence model taking the concatenation of $q$ and $c$ as input and generating the answer, $a$. Since our goal is just to compare the different methods in terms of the quality of the training data generated, we use a T5-small model ($\sim 60M$ parameters), and finetune it with data obtained from each of the considered baselines. We also include a model finetuned on SQuAD~\cite{RajpurkarEtal_2016_Squad_D16-1264}, a widely-used large QA dataset, to illustrate performance when using generic QA data. Since we consider open-ended QA, we evaluate the generated answers using various language generation quality metrics, including BLEU, F1, ROUGE-L, and BLEURT~\cite{SellamEtal_2020_Bleurt_2020.acl-main.704}.

\noindent \textbf{Results:} 
The results summarized in Table~\ref{tab:qa_short} demonstrate the superiority of the data generated with our approach. Our method with only graph-to-text models (i.e. without LLM-based augmentations) outperforms all baselines on \emph{all} metrics. Importantly, when used for question answering, this simplest variant of our method also outperforms baselines where GPT3 was used for data generation. These results suggest that the wide coverage of question types provided by our exhaustive content selection method plays a more significant role than the syntactic diversity of the generated language in the task of question answering. Finally, combining our graph-guided approach for improved coverage with GPT3, for improved language, yields the overall best results by a wide margin.

\paragraph{Question answering with larger models:} Results summarized in Table~\ref{tab:model_abla} show that the generated data provides enough diversity and coverage to support the finetuning of a T5 model with up to 3B parameters, with performance gains consistently improving as a factor of the model's capacity. 
The results also show that smaller T5 models (e.g., T5-base with 0.22B parameters) already provide excellent performance, with the largest variants being competitive against (and even surpassing) the GPT3 and ChatGPT models, despite them being orders of magnitude larger and having been exposed to much larger amounts of data during training (including recipes that likely overlap with our test set).
More interestingly, the FLAN-T5-XL model finetuned on our paraphrased data significantly outperforms GPT models, as well as the variant trained on answer-based augmented data. We attribute this substantial improvement to our proposed question graph transformations, which the LLM answer-based augmentation approach cannot benefit from. These transformations enrich the question pool diversity that models with larger capacity can effectively exploit, resulting in significant performance gain.
More generally, we believe these results underscore the importance of devising approaches to generate domain-specific, high-quality data as proposed in this paper, especially when seeking a more favorable performance-vs-computational cost tradeoff on specific downstream tasks.

\subsection{Intrinsic evaluation}
\label{sec:div-cov}

\begin{table*}[ht]
    \begin{center}
        \small
        \begin{tabular}{ c | l | c  c  c }
        \toprule
         & Generation method & Dist-3 $\uparrow$ & n-gram Div. $\uparrow$ & Coverage $\uparrow$ \\
        \midrule
        \multirow{2}{*}{Rule-based} &Role-based QG \citep{PyatkinEtal_2021_Asking_2021.emnlp-main.108} & 62.3 & 62.7 & 46.5 \\
        &Template-based QG \citep{FabbriEtal_2020_Template_2020.acl-main.413} & 81.1 & 80.4 & 40.5 \\
        \midrule
        \multirow{2}{*}{Learning-based} &Diffusion-based QG \citep{gong2022diffuseq} & 72.4 & 71.5 & 42.8 \\
        &T5-based QG \citep{t5-qg} & 74.7 & 74.3 & 45.6 \\
        \midrule
        \multirow{2}{*}{GPT-based} & GPT3 sentence & 72.7 & 72.7 & 54.9 \\
        & GPT3 recipe & 69.9 & 71.9 & 58.4 \\
        \midrule
        \multirow{3}{*}{Ours} & w/o augmentations & 78.8 & 77.5 & 59.0 \\ 
        & w/ paraphrasing augmentation & 78.3 & 77.6 & 67.3 \\
        & w/ answer-based augmentation & 76.2 & 76.0 & 67.3 \\
        \bottomrule
        \end{tabular}
    \end{center}
    \caption{\textbf{Intrinsic evaluation: question diversity \& coverage.} Comparison of variants of our method and competing approaches in terms of diversity and coverage of the generated questions. Coverage is measured against the human-annotated reference set.}
    \label{tab:q_diver}
\end{table*}

\paragraph{Question diversity and coverage.} We measure the diversity of generated questions in terms of \textit{Dist-n}, the number of distinct n-grams \citep{LiEtal_2016_Diversity_N16-1014}, and \textit{n-gram Diversity}, calculated as $\frac{1}{N} \sum_{n=1}^{N} (\text{Dist-n}), N=5$ \citep{wiher+al.tacl22}.
We compute these metrics from the questions generated using the 70 recipes in the training split.
On the other hand, \textit{coverage} measures how well the human-generated questions in the reference set are covered by questions automatically-generated from the same recipes. The coverage score is defined as $\frac{1}{N^{ref}} \sum_{i=1}^{N^{ref}} {max}_{j \in \tilde{N}} \rho (q^{ref}_i, \tilde{q}_j)$, where $q^{ref}_i$ and $\tilde{q}_j$ denote the $i^{th}$ question in reference set $N^{ref}$ and $j^{th}$ question in generated set $\tilde{N}$, respectively. We use BLEURT metric as the pair-wise scoring function $\rho$, thus the coverage score not only reflects semantic resemblance of generated questions w.r.t. human questions, but also their language naturalness and fluency.

\begin{table*}[ht]
\centering
\small
\begin{tabular}{ l | c c c c c}
    \toprule
    Generation method & Q. Correct & Q. Adeq & Q. Answ & A. Faith & A. Compl \\
    \midrule
    Role-based QG \citep{PyatkinEtal_2021_Asking_2021.emnlp-main.108} & 2.65 & 2.58 & 2.98 & 3.15 & 2.63 \\
    T5-based QG \citep{t5-qg} & 2.95 & 3.10 & 2.63 & 2.73 & 1.68 \\
    GPT3 recipe &  4.70 & 4.60 & 4.30 & 4.43 & 4.15 \\
    \midrule
     Ours w/o augmentations & 3.35 & 3.58 & 3.40 & 3.63 & 2.83 \\
     Ours w/ anwer-based augmentation & 4.73 & 4.60 & 4.55 & 4.43 & 4.08 \\
    \bottomrule
\end{tabular}
\caption{\textbf{Intrinsic evaluation: overall quality via human evaluation.}  Questions are assessed for correctness (Correct), adequacy (Adeq), and answerability (Answ); Answers are assessed for faithfulness (Faith) and completeness (Compl). Scores are up to 5, higher is better.}
\label{tab:q_rating}
\vspace{-10pt}
\end{table*}

\noindent \textbf{Results:} 
Table~\ref{tab:q_diver} shows our intrinsic experimental results and those from competing methods (a summary of all Dist-n scores, $n \in [1,5]$, is included in the appendix). Even when relying solely on a simple graph-to-text generation model (i.e. no LLM augmentations), our method already far exceeds all QG baselines, including GPT3. In terms of diversity, our scores are slightly below those of Template-based QG \citep{FabbriEtal_2020_Template_2020.acl-main.413}. Different from all other methods including ours, in the Template-based QG approach, the questions are generated by shuffling parts of the original sentence to complete templates, therefore retaining most of the original n-gram diversity, however, the synthesized questions lack semantic adequacy and language fluency, reflected by low coverage score. Our method, with or without LLM augmetations, scores substantially better Dist-3 scores than other question generation methods, while also ensuring higher coverage, yielding overall best results. Notably, our method without any augmentation even has slightly better diversity scores than its LLM-augmented variants, indicating that our graph-based content selection approach, which is center amongst three variants, attributes to the richness and exhaustiveness of the synthetic questions. Paraphrasing augmentation, by fixing the language of generated text, further boosts the coverage score by 14\% relative increase. Our variant with answer-based augmentation, although ensuring good coverage, has slightly lower diversity scores, because it lacks the question graph transformations employed by other two variants. In addition to the excellent QA performance of model trained on such data as demonstrated in our extrinsic evaluation, these results signify the quality of data generated by our proposed method.

\paragraph{Overall quality via human evaluation.} 
We further conduct a human study to validate the quality of the generated questions, as well as the match between questions and answers. 
Specifically, we design a human annotation task where the rater assesses the generated questions in terms of grammatical correctness, adequacy, and answerability from the given context. Each entry was rated by 5 different raters, all of them native speakers. 
The corresponding answer is then revealed, and the rater is asked to judge the faithfulness and completeness of the answer. We assess all aspects on a 5-point Likert scale. 
Note that these metrics focus on the \textit{quality} of the questions and answers, which is complementary to the diversity \& coverage metrics.
Details of the human annotation setup and process are in the appendix. 

\noindent \textbf{Results:} 
We perform human study for three baselines, namely, the best rule-based and learning-based methods (according to Table~\ref{tab:q_diver}) and a GPT-based model, in addition to two variants of our method, one with graph-to-text generation, and one with GPT-based question generation. 
Table~\ref{tab:q_rating} shows that data generated with our approach fares well across all annotation aspects compared to other QG baselines. Methods leveraging GPT3, i.e., the ``GPT3 recipe'' and ``Ours w/ answer-based augmentation'', yield the highest quality by a wide margin, highlighting again the complementarity of the two components of our method.

\section{Conclusion}

In this paper we tackled task-specific QA from procedural text. To this end, we proposed a novel method for automatic generation of question-answer pairs from procedural texts in a comprehensive manner, both in their semantic content as well as syntactic diversity. We do so by exploiting the structured nature of the procedural data by using graph-based representations, and devise a systematic way of generating semantically-comprehensive question-answer pairs. We further enrich the syntactic correctness and diversity through the use of LLMs. We show that 1) using automatically-generated in-domain data to train a simple T5 model results in question-answering performance competitive with very large language models such as ChatGPT and GPT3. 2) our method results in excellent coverage of human-generated questions.  

\section{Limitations}

Our proposed method heavily relies on AMRs and Flow Graph representations and thus our method is limited to the few languages supported. Multilingual support may become available once AMR sembank and flow graph corpus are expanded to support multiple languages. Furthermore, errors on the graph-parsing strategies are not mitigated within our method. Finally, we use a simple T5 with standard training to illustrate the performance for question-answering when training with data generated by our method. We believe there is room for further improvements by training more advanced models on our generated data.

\linespread{1.0}


\bibliography{anthology,custom}

\clearpage\newpage

\appendix
\section{Summary}
Here we provide all details. We begin by describing the PQA human data collection (i.e. reference set) and the human study setup in \S \ref{app:data} and \S \ref{app:human-eval}, respectively. Next, we provide more detailed descriptions of the baselines considered in this work in \S \ref{app:QG-baselines}. Finally, we provide further technical details about the proposed approach in \S \ref{app:approach}.



\section{PQA reference dataset collection}\label{app:data}

Recall that we select 100 recipes from BBCfood \citep{bbcfood}, and randomly sample 70 recipes for training and 30 recipes are held off for the test set. For each recipe in the test (reference) set, we set up an interactive cooking simulation with two annotators, where one annotator asks questions that would help complete the task, and the other answers the questions. Both questions and answers are recorded and transcribed afterward. Question annotators are provided with detailed instructions, similar to the example shown in Listing~\ref{lst:data_collect}. On the other hand, answer annotators are given the recipe text and ingredients and asked to interactively give the response to the question annotator. To ensure that QA pairs can be solely answered from the recipe, the answer annotator is instructed to reply with: ``the recipe does not specify that'', anytime a question asked cannot be answered from the recipe text alone.

\begin{listing*}
\centering
\caption{Example simulation instructions to elicit question-answer pairs for a recipe.}
\label{lst:data_collect}
\begin{minted}[frame=single,
               framesep=3mm,
               linenos=true,
               xleftmargin=21pt,
               tabsize=4,
               breaklines,
               fontsize=\footnotesize]{text}
You are asked to cook using the following ingredients:

couscous, chargrilled artichokes, dijon mustard, olive oil, dill leaves, parsley leaves, lemon, watercress, sea bass fillets

There are 8 steps you need to finish.

Your task is to cook by interacting with the system. 

You can ask any questions you have. For example:
    What ingredients do I need in the second step,
    Should I mix ingredient A with ingredient B?,
    Where should I put ...?", "How can I prepare ... ?,
    In step 3, I need to do ..., right?,
    How much ... do I need?,
    Do I need A or B?,
    When should I do ...?
    Why do I need to ...?
    Should I do A first or B first?,
    What ingredients do I need to prepare...?
    ...
The system will provide you with the necessary information. 
You don't have to start with the first step, but to complete the task, you must receive a confirmation for each step of the recipe. 

Note that: 
    * Please imagine that you are in the kitchen, in front of all the ingredients and READY TO COOK.
    * You DO NEED detailed information for cooking, e.g. the order of putting ingredients, the place to put the ingredients, the amount of ingredients you need.
    * Try to ask different types of questions. For example, you are not encouraged to ask "what ingredients do I need for step N?" repetitively.
    * You can only ask general questions, like "what should I do next?", one time throughout the entire process.

\end{minted}
\end{listing*}

\section{Human evaluation details}\label{app:human-eval}

We design a human annotation task to evaluate the overall quality of questions and answers generated by a model. Specifically, we ask human raters to assess a set of generated questions with respect to grammatical correctness, adequacy, and answerability from a given context. For each question, the corresponding answer is then revealed to the rater, and they are asked to judge the faithfulness and completeness of the answer. Each question-answer pair is rated by 5 native English speakers. The human raters are provided with detailed annotation instructions shown in Listing \ref{lst:human study}.

\begin{listing*}
\centering
\caption{Instructions provided to the human raters for assessing the overall quality of question and answer pairs.}
\label{lst:human study}
\begin{minted}[frame=single,
               framesep=3mm,
               linenos=true,
               xleftmargin=21pt,
               tabsize=4,
               breaklines,
               fontsize=\footnotesize]{text}

RECIPE: {RECIPE_TEXT}
QUESTION: {QUESTION}
ANSWER: {ANSWER}

This task contains two phases. In the first phase, you need to read the RECIPE and the QUESTION above. You have to score the
 question on the basis of following three metrics. Note that, when scoring on the basis of one metric, please ignore the re
st two entirely. 

* Grammatical correctness: How well-phrased and grammatical is the question?
    - 1: absolutely grammatically incorrect
    - 2: mostly grammatically incorrect
    - 3: somewhat grammatically incoreect
    - 4: mostly grammatically correct
    - 5: absolutely grammatically correct

* Adequate: Does the question make sense in the context of the recipe?
    - 1: absolutely inadequate
    - 2: mostely inadequate
    - 3: somewhat adequate
    - 4: mostely adequate
    - 5: absolutely adequate

* Answerability: Is it possible to provide an answer to the question ONLY using the information provided in the recipe?
    - 1: absolutely not answerable
    - 2: mostely not answerable
    - 3: somewhat answerable
    - 4: mostely answerable
    - 5: absolutely answerable

Now, if all the numbers about are equal or more than 4, please continue the evaluation below. Otherwise, please click submit.

In the Second phase, you need to continuously read the ANSWER above. You have to score the answer on the basis of following two metrics. Note that, when scoring on the basis of one metric, please ignore the rest two entirely. 

* Faithfulness: Does the answer ONLY contain the information provided in the recipe?
    - 1: absolutely not faithful
    - 2: mostely not faithful
    - 3: somewhat faithful
    - 4: mostely faithful
    - 5: absolutely faithful

* Answer's completeness: Does the provided answer completely address the question?
    - 1: completely fails to address the question
    - 2: mostely fails to address the question
    - 3: somewhat address the question
    - 4: mostely address the question
    - 5: completely address the question
\end{minted}
\end{listing*}

\section{Question generation baselines}\label{app:QG-baselines}
In this section, we provide further technical details for the methods used as baselines.

\paragraph{Rule-based QG} \cite{PyatkinEtal_2021_Asking_2021.emnlp-main.108}. Given a sentence, this method uses rules to generate questions for all semantic roles associated with a given entity, independently of the presence or absence of answers. Since this method is sentence-based, we use it to automatically generate questions for each step in a recipe. Also, since this method does not offer a solution for answer generation, we take the generated questions and accompanying recipe and use Alpaca \cite{alpaca} to automatically generate corresponding answers and filter out questions for which there is no answer in the recipe.

\paragraph{Template-based QG} \cite{FabbriEtal_2020_Template_2020.acl-main.413}. A sentence is segmented into \textit{[Fragment A] + [answer] + [Fragment B]} components. The \textit{[answer]} component is replaced with the question \textit{wh-}word, and re-combined with \textit{[Fragment A]} and \textit{[Fragment B]} in different orders to create questions. Then, similar to the role-based QG baseline, we use Alpaca to generate answers given the generated questions and context.

\paragraph{T5-based QG} \cite{t5-qg}. Following previous work \cite{justask} we use a T5 model \cite{t5} finetuned on the SQuAD dataset \cite{RajpurkarEtal_2016_Squad_D16-1264, t5-qg} for the task of question generation. Specifically, similar to previous work \cite{justask} we provide the finetuned model with the recipe text and let it automatically generate QA pairs. 

\paragraph{Diffusion-based QG} \cite{gong2022diffuseq}. We use a recent approach for non-autoregressive question generation based on text diffusion. We use Alpaca to generate corresponding answers.

\paragraph{GPT3-based QG} \cite{gpt3_neurips20}. We consider GPT3 as an alternative method for question generation. We consider two variants: \textbf{(i)} GPT3-sentence, where we provide GPT3 with each step in the recipe independently and task it with generating \emph{all} possible questions about its content. \textbf{(ii)} GPT3-recipe, where we give the entire recipe text as context and task GPT3 with generating \emph{all} possible questions, with the goal of pushing GPT3 to ask temporal questions that span over multiple steps. 

\section{Intrinsic evaluation}

\begin{table*}[ht]
  \centering
  \scriptsize 
    \begin{tabular}{c|l|ccccccc}
    \toprule
          & \multicolumn{1}{c|}{\textbf{Generation Method}} & \textbf{Dist-1} & \textbf{Dist-2} & \textbf{Dist-3} & \textbf{Dist-4} & \textbf{Dist-5} & \textbf{n-gram Div} & \textbf{Coverage} \\
    \midrule
    \multirow{4}[4]{*}{Rule-based} & Role-based QG \citep{PyatkinEtal_2021_Asking_2021.emnlp-main.108} & 98.7  & 81.1  & 62.3  & 43.9  & 27.5  & 62.7  & 46.5  \\
          & Role-based QG w/ paraphrasing & 98.9  & 83.7  & 67.6  & 51.5  & 36.2  & 67.6 & n/a \\
\cmidrule{2-9}          & Template-based QG \citep{FabbriEtal_2020_Template_2020.acl-main.413} & 95.4  & 90.3  & 81.1  & 72    & 63.2  & 80.4  & 40.5  \\
          & Template-based QG w/ paraphrasing & 97.3  & 89.8  & 79.8  & 69.9  & 60.1  & 79.4 & n/a \\
    \midrule
    \multirow{4}[4]{*}{Learning-based} & Diffusion-based QG \citep{gong2022diffuseq} & 95.2  & 85.9  & 72.4  & 58.7  & 45.4  & 71.5 & 42.8  \\
          & Diffusion-based QG w/ paraphrasing & 97.1  & 87.4  & 75.2  & 62.9  & 50.6  & 74.6 & n/a \\
\cmidrule{2-9}          & T5-based QG \citep{t5-qg} & 96.5  & 87.4  & 74.9  & 62.4  & 50    & 74.2 & 45.6  \\
          & T5-based QG w/ paraphrasing & 98.2  & 88.1  & 76.4  & 64.6  & 52.9  & 76.0 & 50.9  \\
    \midrule
    \multirow{4}[4]{*}{GPT-based} & GPT3 sentence & 99.8  & 86.4  & 72.7  & 59.1  & 45.5  & 72.7  & 54.9  \\
          & GPT3 sentence w/ paraphrasing & 99.2  & 87.7  & 75.4  & 63.1  & 50.9  & 75.3 & 57.8  \\
\cmidrule{2-9}          & GPT3 recipe & 99.6  & 94.9  & 69.9  & 54.9  & 40.1  & 71.9 & 58.4  \\
          & GPT3 recipe w/ paraphrasing & 99.4  & 86.2  & 72.5  & 58.8  & 45.2  & 72.4 & 61.0  \\
    \midrule
    \multirow{3}[2]{*}{Ours} & w/o augmentations & 93.4  & 89.2  & 78.8  & 68.3  & 57.8  & 77.5  & 59.0  \\
          & w/ paraphrasing augmentation & 96.7  & 89.0  & 78.3  & 67.5  & 56.7  & 77.6 & 67.3  \\
          & w/ answer-based augmentation & 98.7  & 88.1  & 76.2  & 64.4  & 52.5  & 76.0 & 67.3  \\
    \bottomrule
    \end{tabular}%
\caption{\textbf{Intrinsic evaluation: question diversity and coverage.} This table shows all scores of all baseline generated datasets and their paraphrased supersets.}
  \label{tab:appendix_intrinsic}%
\end{table*}%

We provide all intrinsic evaluation scores in Table~\ref{tab:appendix_intrinsic}, including all Dist-n metrics, $n \in [1,5]$, as a more complete version of Table 3 in the main paper. We also conduct paraphrasing experiments on all datasets generated by baseline methods. Notably, our method without any augmentation still surpasses the paraphrased supersets of the baselines, showcasing the effectiveness of our proposed graph-based question generation method.

\section{Further technical details}\label{app:approach}

\subsection{Model learning}

\paragraph{Graph-to-question generative model.} While an off-the-shelf AMR-to-text generator \citep{amrlib} works well on general sentences, it often fails to generate correct questions from question AMRs as we observed empirically. This problem may be due to the insufficiency of question data in the standard AMR datasets. To attenuate this issue and improve question generation performance, we finetune a T5-base model to generate questions from AMRs specifically. We parse questions taken from SQuAD \citep{RajpurkarEtal_2016_Squad_D16-1264} and R2VQ \citep{r2vq_paper} datasets into question AMRs and use the data to train our AMR-to-question model. We use the same training setting as AMRlib \citep{amrlib}, secifically we train the T5-base model for 8 epochs using AdamW optimizer, batch size $=8$, starting learning rate $=1e{-4}$ with linear schedule. The model was trained on a single 1080ti GPU in 20hrs.

\paragraph{Question-answering model training.} We train one T5 models for each training dataset (QA pairs generated by one of the QG methods, including baselines and ours) using the same settings as follows: we train each model for 12 epochs using AdamW optimizer with $\beta=\{0.9, 0.999\}$, weight decay $=0.01$, batch size $=256$, starting learning rate $=1e{-5}$ with cosine schedule. T5-small models were trained using eight 1080ti GPUs in under 3hrs. T5-base model was trained in one day using the same GPUs. T5-large and T5-3B/XL were trained for one and two days, respectively, using eight V100 GPUs.

\subsection{GPT3 prompts details}
We use GPT3 for several tasks: (i) as a baseline question generation model, with two variants of GPT3-sentence and GPT3-recipe as explained in \S \ref{app:QG-baselines} above; and
(ii) to augment our approach either as an alternative graph-to-text generation model (GPT3-QG), or as an added component to paraphrase the outputs of our graph-to-text generation module (GPT3-paraphrasing). 
For the GPT3-sentence and GPT3-recipe baselines, we follow up with a prompt to also elicit an answer (Answer Generation).
We describe the prompts used for each case in Table \ref{tab:gpt_prompt}.

\begin{table*}
\centering
\small
\begin{tabular}{l|l}
    \toprule
     Task & Prompt \\
     \midrule
     GPT3-sentence & \makecell[l]{Sentence: \{CONTEXT\} \\ Instruction: Read the above sentence, and ask \{N\_PAIR\} different questions that can only be answered\\ by referring to the given sentence.} \\
     \midrule
     GPT3-recipe & \makecell[l]{Recipe: \{CONTEXT\} \\ Instruction: Read the recipe above, and ask \{N\_PAIR\} different questions that can only be answered \\by referring to the given recipe.} \\
     \midrule
     GPT3-paraphrasing & \makecell[l]{Rewrite this sentence: \{QUESTION\_SENTENCE\}} \\
     \midrule
     GPT3-QG & \makecell[l]{Context: \{CONTEXT\} \\ Instruction: Read the above context, and ask \{N\_PAIR\} different questions that can be answered as\\ ``\{ANSWER\}''. Do not generate answers.} \\
     \midrule
     \midrule
     Answer Generation & \makecell[l]{Answer the question using information in the preceding background paragraph. \\If there is not enough information provided, answer with ``The recipe does not specify''} \\
     \bottomrule
\end{tabular}
\caption{Prompts used to generate questions, answers, or paraphrases for the various GPT3-based models.}
\label{tab:gpt_prompt}
\end{table*}
















\subsection{Question generation from single instrutions}
\label{sec:appendix_single}

\subsubsection{Role-specific QA}

\paragraph{:ARG1} The general algorithm to generate \textit{\textbf{all}} questions on different subgraphs under \textit{:ARG1} is described in Algorithm~\ref{algo:ARG1}.

\noindent\textit{:ARG1 splitting and regrouping.}
One of the limitations of graph-to-text generator is that, if the concept of \textit{:ARG1} is a compound such as in the example of Figure~\ref{fig:amr_q_example} in the main text: ``Stir buttermilk, chili powder and cayenne pepper in a bowl'', then it is unable to generate question about a \textit{:opX} role in the compound (e.g. :op1 (b/ buttermilk):

(s / stir-01 :ARG1 (a / and) :op1 (b / buttermilk) :op2 (p / powder :mod (c / chili)) :op3 (p2 / pepper :mod (c2 / cayenne))) :location (b2 / bowl)) 

Our proposed solution is to transform the above AMR into

(s / stir-01 :ARG1 (b / buttermilk) :ARG2 (a / and) :op1 (p / powder :mod (c / chili)) :op2 (p2 / pepper :mod (c2 / cayenne))) :location (b2 / bowl))

then ask question about \textit{:ARG1} by replacing ``buttermilk'' with \textit{amr-unknown}. We can gradually apply a similar transformation for \textit{:op2} and \textit{:op3} in the above example. 
If \textit{:ARG2} exists in the original sentence, we check whether \textit{:ARG1} and \textit{:ARG2} are semantically equivalent. If they are equivalent, the remaining \textit{:opX} roles in \textit{:ARG1} are merged with \textit{:ARG2}. Otherwise, we convert the original \textit{:ARG2} role into \textit{:instrument} or \textit{:location}, and then split \textit{:ARG1}.

\begin{algorithm*}
\caption{Generate questions about \textit{:ARG1}}
\label{algo:ARG1}
\begin{algorithmic}[1]
\Procedure{generate\_ARG1\_attribute\_question}{}
\State ret = \{\}
\For{role in :ARG1 concept sub-roles}
\If {role = :mod}
\If {exist\_role(:quant)}:
\State remove\_role(:quant)
\EndIf
\State ret.add(replace\_amr\_unknown(:mod))
\ElsIf {role = :quant}
\State ret.add(replace\_amr\_unknown(:quant))
\EndIf
\EndFor
\State return ret
\EndProcedure
\end{algorithmic}

\begin{algorithmic}[1]
\Procedure{generate\_ARG1\_questions}{}
\State ret = \{\}
\State ret.add(replace\_amr\_unknown(:ARG1))
\If {:ARG1 concept is single entity}
\State ret.add(generate\_ARG1\_attribute\_question())
\ElsIf {:ARG1 concept is compound}
\State Split entities in :ARG1
\For {each entity}
\State Set entity as concept of :ARG1, other entities form :ARG2
\State ret.add(replace\_amr\_unknown(:ARG1))
\State ret.add(generate\_ARG1\_attribute\_question())
\EndFor
\EndIf
\State return ret
\EndProcedure
\end{algorithmic}
\end{algorithm*}

\paragraph{:ARG2} Directly placing \textit{amr-unknown} on \textit{:ARG2} would not work most of the time. We empirically found that, the graph-to-text generator is unable to generate correct question if the \textit{amr-unknown} concept is placed directly on \textit{:ARG2} role. This limitation may originate from AMR data that the model was trained on, which does not contain questions on \textit{:ARG2}. Furthermore, in order to prepare question data to finetune the generator, we used the text-to-graph parser to parse questions in the R2VQ dataset into question AMRs, from which we observed that there was not any questions on \textit{:ARG2}. 
Thus, we proposed a solution that swaps \textit{:ARG1} and \textit{:ARG2}, turning \textit{:ARG2} into \textit{:ARG1}, from which we can generate questions about \textit{:ARG2} (now in the form of \textit{:ARG1}). However, there are two major problems with swapping: 1) Whether the concept of \textit{:ARG2} is a food item. If \textit{:ARG2} describes a tool then swapping will invalidate the original sentence.
2) Whether the concepts of \textit{:ARG1} and \textit{:ARG2} are swappable - in other words, are they of equivalent roles?. For example, the sentence ``Mix chicken with spices'' and its transformed version ``Mix spices with chicken'' are semantically equivalent, but ``Add spices to chicken” and “Add chicken to spices'' are not.

To address the first problem, we create a filter to check if \textit{:ARG2} is a tool or not. We do so by first gathering all \textit{:instrument} concepts from the YouCook2 dataset, and during QA generation, we check if the concept of \textit{:ARG2} is an instrument among the list, in that case we convert \textit{:ARG2} core role to :instrument role and ask question on \textit{:instrument} instead. 
To solve the second problem, we devise a set of rules to determine if \textit{:ARG2} and \textit{:ARG1} are semantically equivalent. Firstly, we check the verb if it implies moving direction or not. Such verbs include {``add'', ``put'', ``pour'', etc.}. Secondly, because \textit{:ARG2} typically follows a preposition, we check if the preposition is directional, ie. it’s among {``in'', ``on'', ``to'', ``into'', ``over''}. In such cases, we do not carry out swapping and instead convert \textit{:ARG2} into \textit{:location}, and ask question about :location as usual.

One example is shown in Figure \ref{fig:ARG2}.

\begin{figure*}
\centering
\begin{boxedminipage}{25em}
- Original sentence:
\\\textcolor{red}{We mix salt and chicken.}
\\(m / mix-01
\\\hspace*{1em}      :ARG0 (w / we)
\\\hspace*{1em}      :ARG1 (s / salt)
\\\hspace*{1em}      :ARG2 (c / chicken))
\\- Directly replace concept of ARG2 with amr-unknown:
\\(m / mix-01
\\\hspace*{1em}      :ARG0 (w / we)
\\\hspace*{1em}      :ARG1 (s / salt)
\\\hspace*{1em}      :ARG2 (a / amr-unknown))
\\\textcolor{blue}{How much salt do we mix?}
\\- Swap concepts of ARG1 and AGR2:
\\(m / mix-01
\\\hspace*{1em}      :ARG0 (w / we)
\\\hspace*{1em}      :ARG1 (a / amr-unknown)
\\\hspace*{1em}      :ARG2 (s / salt))
\\\textcolor{blue}{What do we mix with salt?}
\end{boxedminipage}
\caption{Example questions generated for the concept of \textit{:ARG2} role.}
\label{fig:ARG2}
\end{figure*}

\paragraph{:time} The procedure is shown in Algorithm~\ref{algo:time}. Notably, in order to overcome the limitation of the AMR-to-text generator, we first remove all non-core roles from the AMR graph, except for \textit{:time}.

\begin{algorithm}
\caption{Gennerate questions about \textit{:time}}
\label{algo:time}
\begin{algorithmic}[1]
\State Remove all roles except :ARG1, :ARG2, :time
\If {concept of :time starts with ``until''}
\State replace :time with :extent
\State return replace\_amr\_unknown(:extent)
\Else
\State return replace\_amr\_unknown(:time)
\EndIf
\end{algorithmic}
\end{algorithm}

\paragraph{Quantity (-quantity) concepts.} This section applies to all quantity concepts, except \textit{temporal-quantity} which often appears in :duration role. The procedure is shown in Algorithm~\ref{algo:quant}. The key idea is to search for the \textit{:quant} role within the subgraph of \textit{-quantity} concept, and replace its concept (or rather, value) with \textit{amr-unknown}. Note that, due to limitations of the text generator, we are unable to generate the correct question on ``:quant'' in a large graph, hence before that, we must simplify the graph.

\begin{algorithm}
\caption{Generate questions about \textit{quantity}.}
\label{algo:quant}
\begin{algorithmic}[1]
\State Remove all roles except :ARG1, :ARG2, :location, and the role in question.
\State return replace\_amr\_unknown(:quant)
\end{algorithmic}
\end{algorithm}

\paragraph{Other roles} The other supported roles are: \textit{:duration}, \textit{:location}, \textit{:instrument}, \textit{:mod}, \textit{:domain}, \textit{:purpose}, \textit{:accompanier}, \textit{:degree}, \textit{:value} and \textit{:quant}. Their questions are generated as described in Algorithm~\ref{algo:direct_question}. With a few exceptions, we can simply replace the concept of a target role with \textit{amr-unknown} to generate a question about that role.

\begin{algorithm}
\caption{Direct question generation for AMR roles.}
\label{algo:direct_question}
\begin{algorithmic}[1]
\If {role = :mod}
\State remove\_role(:quant)
\State return replace\_amr\_unknown(:mod)
\ElsIf {role = :quant}
\State Remove all roles except :ARG1, :ARG2, :location, and the target role.
\State return replace\_amr\_unknown(:quant)
\Else
\State return replace\_amr\_unknown(role)
\EndIf
\end{algorithmic}
\end{algorithm}

\subsubsection{Instruction-level questions}

There are two types of questions to ask \textit{``how to do something''}:
\begin{itemize}
    \item How do you [do something]?
    \item What do we do with [something]?
\end{itemize}

\noindent\textbf{How do you [do something]?} The procedure is described in Algorithm~\ref{algo:how}. The goal is achieved by adding the \textit{:manner} role with \textit{amr-unknown} concept. To overcome the limitation of AMR-to-text generation, we remove all non-core roles from the AMR graph before generating the question.

\begin{algorithm}
\caption{\textit{``How''} question generation}
\label{algo:how}
\begin{algorithmic}[1]
\State Remove all non-ARG roles and corresponding concepts
\State return add\_role(:manner, amr-unknown)
\end{algorithmic}
\end{algorithm}

\noindent\textbf{What do we do with [something]?} We generate questions for the whole set of original entities in \textit{:ARGx}, as well as every single entity. The procedure is described in Algorithm~\ref{algo:what_do_with}, and is summarized here:
\begin{itemize}
    \item Grouping all \textit{:ARGx} into \textit{:ARG2}. This step basically combines all food items into one single compound defining \textit{``something''}.
    \item Adding \textit{:ARG1} with \textit{amr-unknown} concept, to enable question generation.
    \item Replacing the current verb frame with \textit{``do-02''} frame.
\end{itemize}

\begin{algorithm}
\caption{\textit{``What do we do with [something]?''} question generation}
\label{algo:what_do_with}
\begin{algorithmic}[1]
\State ret = \{\}
\State Remove all non-ARG roles and corresponding concepts
\If {:ARG2 exists}
\State merge all :ARGx into :ARG1
\EndIf
\State replace\_concept(graph\_top, ``do-02'')
\State Rename :ARG1 $\rightarrow$ :ARG2
\State //generate one question about the entire compound in :ARG2
\State ret.add(add\_role(:ARG1, amr-unknown)
\State //generate question about every single entity in :ARG2
\State Split entities in :ARG2
\For {each entity}
\State Set entity as the sole concept of :ARG2 $\rightarrow$ new\_graph 
\State ret.add(new\_graph)
\EndFor
\State return ret
\end{algorithmic}
\end{algorithm}

Some examples are shown in Listing~\ref{lst:what_do_we_do}.

\begin{listing*}
\centering
\caption{Examples of generating ``What do we do...?'' questions.}
\label{lst:what_do_we_do}
\begin{minted}[frame=single,
               framesep=3mm,
               linenos=true,
               xleftmargin=21pt,
               tabsize=4,
               fontsize=\footnotesize]{text}
- The original instruction: 
"Fry the coated chicken wings in oil at 350 degrees for 3-5 mins."

- Original graph:
(f / fry-01
      :mode imperative
      :ARG0 (y / you)
      :ARG1 (w / wing
            :part-of (c / chicken)
            :ARG1-of (c2 / coat-01))
      :ARG2 (o / oil)
      :location (t / temperature-quantity
            :quant 350
            :scale (c3 / celsius))
      :duration (b / between
            :op1 (t2 / temporal-quantity
                  :quant 3
                  :unit (m / minute))
            :op2 (t3 / temporal-quantity
                  :quant 5
                  :unit (m2 / minute))))

- Simplifying the graph:
(f / fry-01
      :mode imperative
      :ARG0 (y / you)
      :ARG1 (w / wing
            :part-of (c / chicken)
            :ARG1-of (c2 / coat-01))
      :ARG2 (o / oil))

- Question on all food items:
(f / do-02
    :ARG0 (y / you)
    :ARG2 (a / amr-unknown
        / and
        :op1 (w / wing
            :part-of (c / chicken)
            :ARG1-of (c2 / coat-01))
        :op2 (o / oil))
    :ARG1 a)
What do you do with a coated chicken wing and oil?

- Question on single entity:
(f / do-02
    :ARG0 (y / you)
    :ARG2 (w / wing
        :part-of (c / chicken)
        :ARG1-of (c2 / coat-01))
    :ARG1 (a / amr-unknown))
What do you do with a chicken's coated wings?
--------------------------------
(f / do-02
    :ARG0 (y / you)
    :ARG2 (o / oil)
    :ARG1 (a / amr-unknown))
What do you do with oil?
\end{minted}
\end{listing*}

\subsubsection{Polarity ``yes/no'' questions} The procedure is described in Algorithm~\ref{algo:yes_no}. The key idea is to add a new node to the original AMR, with the concept of \textit{amr-unknown} connected to the main verb node with the \textit{:polarity} role. 

\begin{algorithm*}
\caption{``Yes/No'' question generation}
\label{algo:yes_no}
\begin{algorithmic}[1]
\State remove\_role(:mode)
\State add\_role(:ARG0, choice(\{``I'',``we'',``you''\}))
\State add\_role(:polarity, amr-unknown)
\State sample\_and\_replace(role) for role in orginal\_AMR       \# for ``No'' question
\end{algorithmic}
\end{algorithm*}

\subsection{Temporal question generation}

\subsubsection{Instructions \& action graph}

\begin{figure*}
    \centering
    \includegraphics[width=\textwidth]{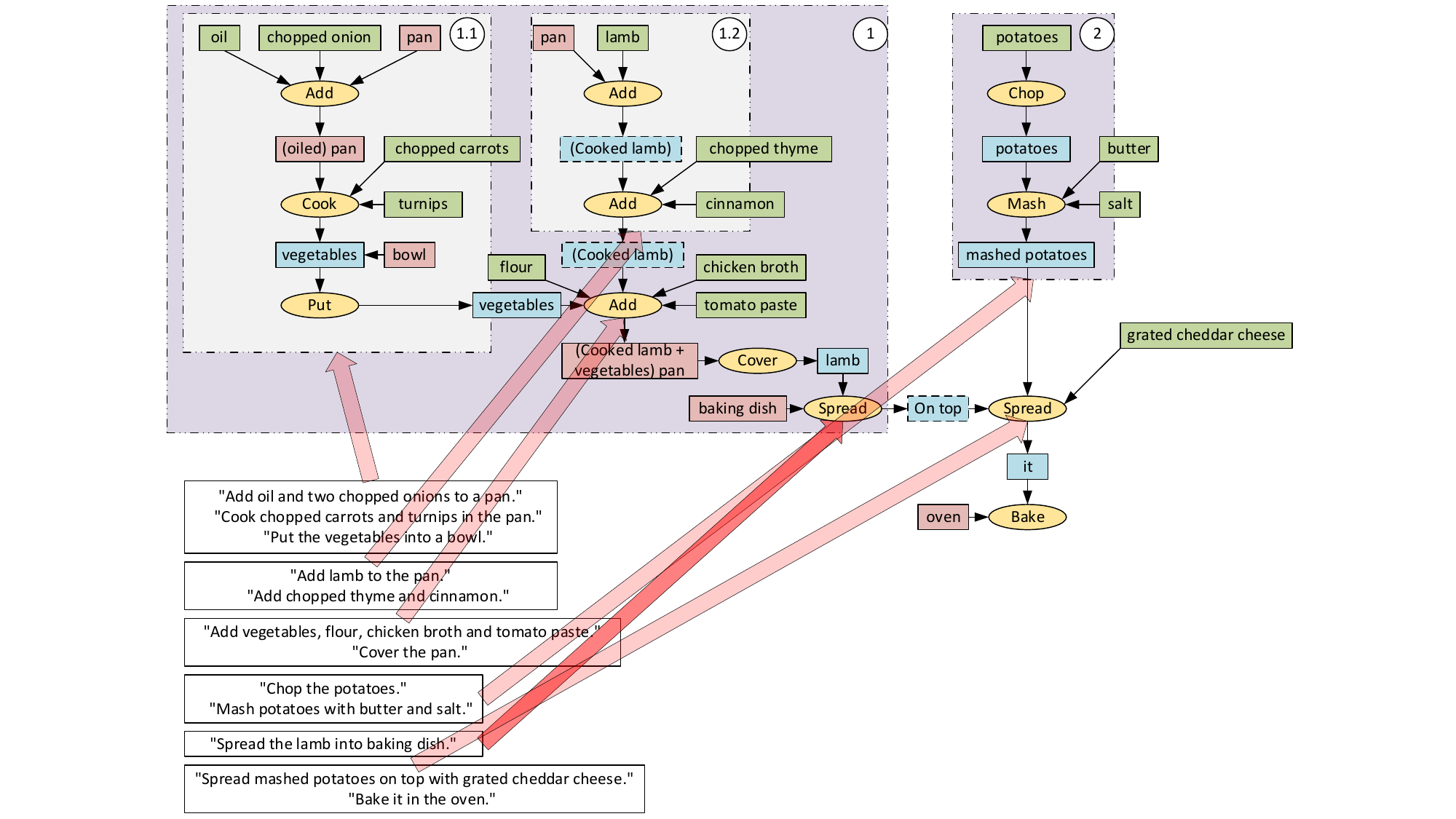}
    \caption{An example of a cooking recipe (divided into subtasks, each containing several instructions), and the corresponding flow graph (divided into subgraphs corresponding to each subtask). We can see that the recipe may be followed in a different order than the sequential ordering of the steps in the written recipe. 
     }
    \label{fig:inst_graph}
\end{figure*}

Figure~\ref{fig:inst_graph} shows an example of a cooking recipe  and its corresponding flow graph. As can be seen in the flow graph, the dependencies among actions and other cooking entities (e.g., ingredients and intermediate food items) do not necessarily follow the sequential order of the steps in the recipe.

\subsubsection{Temporal question templates and examples}

\paragraph{Composition of mixture.} We design 12 question templates, listed below:

\begin{easylist}[enumerate]
& ``What are the ingredients of the \textbf{\{mixture\_name\}}?''
& ``What are the ingredients to prepare the \textbf{\{mixture\_name\}}?''
& ``What are the ingredients required for the \textbf{\{mixture\_name\}}?''
& ``What are the ingredients required to prepare the \textbf{\{mixture\_name\}}?''
& ``What are the ingredients needed for the \textbf{\{mixture\_name\}}?''
& ``What are the ingredients needed to prepare the \textbf{\{mixture\_name\}}?''
& ``What is in the \textbf{\{mixture\_name\}}?''
& ``What ingredients are in the \textbf{\{mixture\_name\}}?''
& ``What ingredients go into the \textbf{\{mixture\_name\}}?''
& ``What ingredients are for the \textbf{\{mixture\_name\}}?''
& ``What ingredients make the \textbf{\{mixture\_name\}}?''
& ``What do I need for the \textbf{\{mixture\_name\}}?''
\end{easylist}

We only apply these question templates with a \textbf{\textit{named mixture}}, and ignore implicit mixtures and pronouns (such as ``it'' and ``them''). The procedure is described in Algorithm~\ref{algo:graph_mixture}. Some examples of questions generated from the graph in Figure~\ref{fig:inst_graph} are shown in Listing~\ref{lst:graph_mixture}.

\begin{algorithm*}
\caption{Generate questions about ``mixture''.}
\label{algo:graph_mixture}
\begin{algorithmic}[1]
\Procedure{Get\_ingr\_of\_mixture}{graph, mixture}
\State $prev\_act\_id = graph[mixture].prev\_act\_id$
\State $action = graph[prev\_act\_id]$
\State $ret = \{\}$
\For{$ingr \in action.input$}
\If{$action.input[ingr] < 0$}
\State ret.add(ingr)
\Else
\State $others = get\_ingr\_of\_mixture(graph, ingr)$
\If {$len(others) > 0$}
\State ret.add(others)
\EndIf
\EndIf
\EndFor
\State return ret
\EndProcedure
\end{algorithmic}

\begin{algorithmic}[1]
\Procedure{Generate\_mixture\_question}{graph, templates}
\State $ret = \{\}$
\For {$action \in graph.action\_with\_mixtures()$}
\For {$mixture \in action.mixtures()$}
\State $ingrs = get\_ingr\_of\_mixture(graph, mixture)$
\State $answer = create\_answer(ingrs)$
\For {$template \in templates$}
\State $question = create\_question(template, mixture)$
\State ret.add((question, answer))
\EndFor
\EndFor
\EndFor
\State return ret
\EndProcedure
\end{algorithmic}
\end{algorithm*}

\begin{listing*}
\centering
\caption{Examples of generating questions about a ``mixture''.}
\label{lst:graph_mixture}
\begin{minted}[frame=single,
               framesep=3mm,
               linenos=true,
               xleftmargin=21pt,
               tabsize=4,
               fontsize=\footnotesize]{text}
The original instruction: 
"Put the vegetables into a bowl."

Q: What is the ingredient in vegetable preparation? (type 2)
(ii / ingredient
    :domain (a / amr-unknown)
    :purpose (p / prepare-01
        :ARG1 (v / vegetable)))

Q: What ingredients are required to prepare vegetables? (type 4)
(r / require-01
    :ARG1 (ii / ingredient
        :domain (a / amr-unknown))
    :purpose (p / prepare-01
        :ARG1 (v / vegetable)))


A: Chopped carrots and turnips.
(c / chop-01
    :ARG1 (a / and
        :op1 (c2 / carrot)
        :op2 (t / turnip)))
\end{minted}
\end{listing*}

\paragraph{Questions about preceding/next action.} In this task we employ two templates:
\begin{itemize}
    \item ``What do we do before $A_i$?''
    \item ``What do we do after $A_i$?''.
\end{itemize}

The algorithm to generate ``next'' action is given in Algorithm~\ref{algo:graph_next_act}. Notice in this algorithm, we limit $A_k$ to those with $k > i$. Some examples are shown in Listing~\ref{lst:graph_next_action} and~\ref{lst:graph_prev_action}. To generate ``before'' question, we will find the previous action instead of the next one in the flow graph.

\begin{algorithm*}
\caption{Generate questions about the next action.}
\label{algo:graph_next_act}
\begin{algorithmic}[1]
\Procedure{Generate\_next\_action\_question}{graph, templates}
\State $ret = \{\}$
\For {$action \in graph.actions()$}
\State $next\_actions = \{\}$
\If {$action.next\_action \ne NULL$}
\State next\_actions.add(action.next\_action)
\State $other\_actions = find\_prev\_actions(graph, action.next\_action)$
\For{$a \in other_actions$}
\If{$a.id > action.id$}
\State next\_actions.add(a)
\EndIf
\EndFor
\EndIf
\If {$len(next\_actions) > 0$}
\State questions = create\_question(templates, action)
\For{$a \in next\_actions$}
\State answer = get\_action(graph, a)
\For {$question \in questions$}
\State ret.add((question, answer))
\EndFor
\EndFor
\EndIf
\EndFor
\State return ret
\EndProcedure
\end{algorithmic}
\end{algorithm*}

\begin{listing*}
\centering
\caption{Examples of generating questions about the next action (from recipe in Fig.~\ref{fig:inst_graph} above).}
\label{lst:graph_next_action}
\begin{minted}[frame=single,
               framesep=3mm,
               linenos=true,
               xleftmargin=21pt,
               tabsize=4,
               fontsize=\footnotesize]{text}
The instruction in focus (#7): 
"Chop the potatoes."

Q: What will we do next?
(d / do-02
    :ARG0 (w / we)
    :ARG1 (a / amr-unknown)
    :time (n / next))

Q: What do we do after chopping potatoes?
(d / do-02
    :ARG0 (w / we)
    :ARG1 (a / amr-unknown)
    :time (a2 / after
        :op1 (c / chop-01
            :ARG1 (p / potato))))

A: Mash potatoes with butter and salt.
(m / mash-01
    :mode imperative
    :ARG0 (y8 / you)
    :ARG1 (p8 / potato)
    :accompanier (a10 / and
        :op1 (b4 / butter)
        :op2 (s / salt)))
\end{minted}
\end{listing*}

\begin{listing*}
\centering
\caption{Examples of generating questions about preceding action (from recipe in Fig.~\ref{fig:inst_graph} above).}
\label{lst:graph_prev_action}
\begin{minted}[frame=single,
               framesep=3mm,
               linenos=true,
               xleftmargin=21pt,
               tabsize=4,
               fontsize=\footnotesize]{text}
The instruction in focus (#10): 
"Spread mashed potatoes on top with grated cheddar cheese."

Q: What do we do before spreading mash potatoes on top with grated cheddar cheese?
(d / do-02
    :ARG0 (w / we)
    :ARG1 (a / amr-unknown)
    :time (b / before
        :op1 (s / spread-01
            :ARG1 (p / potato
                :ARG1-of (m / mash-01))
            :ARG2 (t / top)
            :accompanier (c / cheese
                :mod c
                :mod (c2 / cheddar))
            :ARG1-of (g / grate-02))))
            
A: Mash potatoes with butter and salt.
\end{minted}
\end{listing*}

\paragraph{Questions about the order of actions.} We adopt two templates:
\begin{itemize}
    \item ``Do we do A or do we do B first?'': using AMR ``or'' composition frame.
    \item ``Doing A and doing B, which is first?'': using AMR ``amr-choice'' composition frame.
\end{itemize}

We also swap A \& B, so for each pair of $\{A_i, A_j\}$ we can generate four questions. Full examples are shown in Listing~\ref{lst:graph_action_order}.

\begin{listing*}
\centering
\caption{Examples of generating ``which is first'' questions.}
\label{lst:graph_action_order}
\begin{minted}[frame=single,
               framesep=3mm,
               linenos=true,
               xleftmargin=21pt,
               tabsize=4,
               fontsize=\footnotesize]{text}
Instruction #0: Add oil and two chopped onions to a pan.
Instruction #1: Cook chopped carrots and turnips in the pan.

Question 1: "First, do we add oil and 2 chopped onions to the pan, 
             or do we cook the chopped carrots and turnips in the pan?"
(o3 / or
    :op1 (a / add-02
        :ARG0 (w / we)
        :ARG1 (a2 / and
            :op1 (o / oil)
            :op2 (o2 / onion
                :quant 2
                :ARG1-of (c / chop-01)))
        :ARG2 (p / pan))
    :op2 (c4 / cook-01
        :ARG1 (a3 / and
            :op1 (c2 / carrot
                :ARG1-of (c3 / chop-03))
            :op2 (t / turnip))
        :location (p2 / pan)
        :ARG0 w)
    :polarity (a4 / amr-unknown)
    :ord (o4 / ordinal-entity
        :value 1))

Question 2: "First, add oil and 2 chopped onions to the pan, 
             or cook the chopped carrots and turnip in the pan?"
(a / amr-choice
    :op1 (a3 / add-02
        :ARG1 (a2 / and
            :op1 (o / oil)
            :op2 (o2 / onion
                :quant 2
                :ARG1-of (c / chop-01)))
        :ARG2 (p / pan))
    :op2 (c4 / cook-01
        :ARG1 (a4 / and
            :op1 (c2 / carrot
                :ARG1-of (c3 / chop-03))
            :op2 (t / turnip))
        :location (p2 / pan))
    :ord (o3 / ordinal-entity
        :value 1))

Answer: "First, add oil and 2 chopped onions to the pan."
(a3 / add-02
    :ARG1 (a2 / and
        :op1 (o / oil)
        :op2 (o2 / onion
            :quant 2
            :ARG1-of (c / chop-01)))
    :ARG2 (p / pan)
    :ord (o3 / ordinal-entity
        :value 1))
\end{minted}
\end{listing*}



\end{document}